\documentclass[sigconf, nonacm]{acmart}
\usepackage{tabularx}
\usepackage{pifont}

\newcommand{\eg}{\emph{e.g.}}
\newcommand{\ie}{\emph{i.e.}}

\usepackage{wasysym}
\newcommand{\cmark}{\CIRCLE}       
\newcommand{\pmark}{\RIGHTcircle}  
\newcommand{\xmark}{\Circle}       
\usepackage{tcolorbox}
\usepackage[table]{xcolor}
\usepackage{url}
\usepackage{multirow}
\usepackage{subcaption}
\usepackage[linesnumbered,ruled,vlined]{algorithm2e}

\AtBeginDocument{%
  }

\newcommand{\dataset}{\small{\textsc{{XNote}}}}

\DeclareTextFontCommand{\datafield}{\ttfamily\small}

\begin{document}



\title{\textsc{XNote}: Benchmarking Automated Community Notes Generation for Image-based Contextual Deception}


\author{
Jin Ma, Jingwen Yan, Mohammed Aldeen, Ethan Anderson, Taran Kavuru, \\
Jinkyung Katie Park, Feng Luo, Long Cheng
}
\affiliation{%
  \institution{School of Computing, Clemson University}
  \state{South Carolina}
  \country{USA}
}
\renewcommand{\shortauthors}{Jin Ma, et al.}

\begin{abstract}
Community Notes have emerged as an effective crowd-sourced mechanism for combating online deception on social media platforms. However, its reliance on human contributors limits both the timeliness and scalability. In this work, we study the automated Community Notes generation task for \textit{image-based contextual deception}, where an authentic image is paired with misleading context (\eg, time, entity, and event). Unlike prior work that primarily focuses on deception detection (\ie, judging whether a post is true or false in a binary manner), automated Community Notes generation requires producing concise and grounded notes that help users recover the missing or corrected context. This problem remains underexplored due to the scarcity of datasets that support this task. To address this gap, we curate a real-world dataset, {\dataset}, comprising X posts with associated Community Notes and external contexts, along with annotations of topics and deceptive factors. We further benchmark a range of frontier large vision language models (LVLMs) on {\dataset}, evaluating their performance on both deception detection and note generation tasks. We also compare against an end-to-end approach, SNIFFER~\cite{qi2024sniffer}, and a commercial tool, GPT-5. Our results highlight the challenges in automated Community Notes generation, underscoring the need for improved methods and metrics tailored for this task.

\end{abstract} 

\keywords{Community Notes generation, contextual deception}


\maketitle

\section{Introduction}
\begin{figure}
    \centering
    \includegraphics[width=\linewidth]{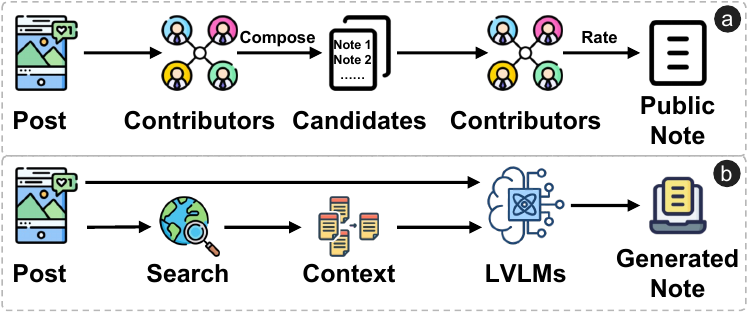}
    \caption{(a) Workflow by which a Community Note becomes publicly available. (b) Pipeline of automated Community Notes generation with web search and LVLMs.}
    \label{fig:xcomm_example}
\end{figure}

In recent years, the rapid spread of online deception\footnotemark on social media platforms has emerged as a significant threat to social trust and public health~\cite{tsikerdekis2014online, pennycook2021psychology, lewandowsky2023misinformation, aimeur2023fake}. For instance, misinformation surrounding the COVID-19 vaccine has led to hesitancy and undermined public health efforts during the pandemic period~\cite{aw2021covid,allen2024quantifying}. To curb the spread of online deception and build responsible online social networks, major social platforms have started to flag suspicious content by overlaying reminder text or adding extra context. On X (formerly Twitter), Community Notes offer a crowd-sourced fact-checking approach~\cite{xcommunitynote} that allows registered contributors to add context-corrective notes to deceptive posts.\footnotetext{{Here the online deception indicates that deceivers use social media platforms to deliver false beliefs to audiences~\cite{tsikerdekis2014online}.}}
Meta has also introduced a similar community note feature on Facebook and Instagram~\cite{metacommunitynote}, as announced in March 2025. Figure~\ref{fig:xcomm_example}(a) illustrates the typical Community Notes workflow: registered contributors first draft candidate notes, other contributors then rate them, and a note is published only after receiving sufficient approval. However, this pipeline depends on broad and sustained human participation in both writing and rating, introducing latency and constraining responsiveness at scale. These limitations underscore the need for automated approaches that can generate notes in a timely manner while reducing manual effort.


Recent studies have shown that on social media platforms, image-based posts attract higher user engagement and can be more persuasive than text-only deception~\cite{yang2023visual,dufour2024ammeba,li2020picture,wang2021understanding}. Consequently, we focus on image-based deception, which is commonly launched by attackers through two mechanisms~\cite{dufour2024ammeba}: (i) \textit{content} manipulation, where the image itself is fabricated or altered, \ie, via Photoshop~\cite{photoshop} or deepfake tools~\cite{westerlund2019emergence}; and (ii) \textit{context} manipulation (also known as out-of-context, OOC), where an authentic image is paired with misleading text. These two types of deception require different mitigation strategies. Content manipulation is typically addressed with pixel-level forensics that target AI artifacts or fingerprints~\cite{mirsky2021creation,lin2024detecting}, while context manipulation is addressed by retrieving external evidence to determine the correct context~\cite{qi2024sniffer,yuan2023support}. Our work targets \textit{context} manipulation (\ie, contextual deception) because it naturally aligns with the goals of Community Notes. Specifically, X's Community Notes guide~\cite{xcommunitynote} emphasizes that helpful notes should satisfy five criteria: (1) credibility: cite high-quality sources; (2) clarity: be easy to understand; (3) relevance: directly address the post’s claim; (4) veracity: provide important context; and (5) neutrality: use neutral or unbiased language. Meeting these criteria depends primarily on grounding claims in credible external evidence and clearly conveying corrective context~\cite{de2025supernotes,solovev2025references}, rather than detecting pixel-level manipulation within the image itself.


Despite extensive research on automated online deception detection, \ie, determining whether the information is true or false~\cite{abdali2024multi,hu2025overview}, little effort has been put into Community Notes generation. This gap largely stems from the scarcity of datasets: most existing image-based deception datasets are either synthetic and fail to capture real-world complexity~\cite{luo2021newsclippings,xu2024mmooc,liu2024mmfakebench}, are designed primarily for detection without explanatory notes~\cite{nakamura2019fakeddit,papadopoulos2024verite}, or provide rationales that are not directly usable in a Community Notes-style format~\cite{rangapur2025fin, dufour2024ammeba}. To address this challenge, we curate a real-world dataset, {\dataset}, consisting of X posts, their corresponding Community Notes, external context retrieved via reverse web search, along with annotations of post topics and deceptive factors. We then benchmark a range of frontier large vision language models (LVLMs) on both deception detection and note generation tasks under two settings: (i) a \textit{closed-book} setting, where models rely solely on their internal knowledge, and (ii) a \textit{naive RAG} (retrieval-augmented generation) pipeline that incorporates web search results as additional context, as shown in Figure~\ref{fig:xcomm_example}(b). Our contributions are summarized as follows:
\begin{itemize}
\item We curate {\dataset}, a real-world dataset for automated Community Notes generation, containing X posts, ground-truth Community Notes, external context, and detailed annotations of topics and manipulation factors.
\item We perform comprehensive evaluations on both deception detection and note generation tasks, comparing multiple frontier LVLMs under closed-book and retrieval-augmented settings, as well as an end-to-end approach (SNIFFER~\cite{qi2024sniffer}) and a commercial system (GPT-5~\cite{gpt}).
\end{itemize}

\begin{table}[t]
\centering
\caption{Comparison of {\dataset} with existing contextual deception datasets. Symbols denote availability: \xmark (no), \cmark (yes), \pmark (partial). ``Fact-checked'' indicates whether data originates from professional fact-checking sources.}
\label{tab:dataset_comparison}
\small
\resizebox{\linewidth}{!}{
\begin{tabular}{lcccc}
\toprule
\textbf{Dataset} & \textbf{Source} & \textbf{Explanations} & \textbf{Fact-checked} & \textbf{Note-style} \\
\midrule
NewsCLIPpings~\cite{luo2021newsclippings} & Synthetic & \xmark & \xmark & \xmark \\
MMOOC~\cite{xu2024mmooc} & Synthetic & \xmark & \xmark & \xmark \\
MMFakeBench~\cite{liu2024mmfakebench} & Synthetic & \xmark & \xmark & \xmark \\
Fakeddit~\cite{nakamura2019fakeddit} & Real-world & \xmark & \xmark & \xmark \\
VERITE~\cite{papadopoulos2024verite} & Real-world & \xmark & \cmark & \xmark \\
AMMeBa~\cite{dufour2024ammeba} & Real-world & \cmark & \cmark & \xmark \\
Fin-Fact~\cite{rangapur2025fin} & Real-world & \cmark & \cmark & \xmark \\
\midrule
{\dataset} (Ours) & Real-world & \cmark & \pmark & \cmark \\
\bottomrule
\end{tabular}
}
\end{table}

\section{Related Work}
In the field of image-based contextual deception, several datasets synthesize deceptive examples from non-deceptive data. For instance, NewsCLIPpings~\cite{luo2021newsclippings} begins with genuine image-text pairs from news outlets and constructs context manipulation cases by semantically mismatching images and captions using CLIP~\cite{radford2021learning} and SBERT~\cite{reimers2019sentence}. Similarly, MMOOC~\cite{xu2024mmooc} recombines verified news articles to produce synthetic mismatches. MMFakeBench~\cite{liu2024mmfakebench} instead leverages generative models and AI tools to construct a mixed-source multimodal misinformation benchmark. While such synthesization enables large-scale data creation, it may not capture the full complexity of real-world deceptive content on social media platforms. By contrast, Fakeddit~\cite{nakamura2019fakeddit} and VERITE~\cite{papadopoulos2024verite} collect  real-world data from online websites, \eg, Reddit and professional fact-checkers, but provide only veracity labels, without explanatory context. Recent datasets begin to include explanations to support further research. AMMeBa~\cite{dufour2024ammeba} compiles a large-scale dataset of fact-checked claims with associated media through ClaimReview~\cite{schemaorg_claimreview,google_claimreview_structured_data}. Fin-Fact~\cite{rangapur2025fin} aggregates image-text items and explanations from sources such as PolitiFact~\cite{politifact}, Snopes~\cite{snopes}, and FactCheck~\cite{factcheck} with a focus on finance. However, these datasets are primarily drawn from already fact-checked items, limiting coverage of in-the-wild deception. 
Therefore, {\dataset} directly collects data from the X social media platform, covering data that has not been fact-checked yet (Section~\ref{sec:analysis}). Our {\dataset} further provides note-style ground-truth annotations that directly support automated Community Notes generation. Table~\ref{tab:dataset_comparison} summarizes the key differences between {\dataset} and existing datasets.

\section{Dataset Collection and Analysis}

\subsection{Dataset Collection} 
\begin{figure}
    \centering
    \includegraphics[width=1\linewidth]{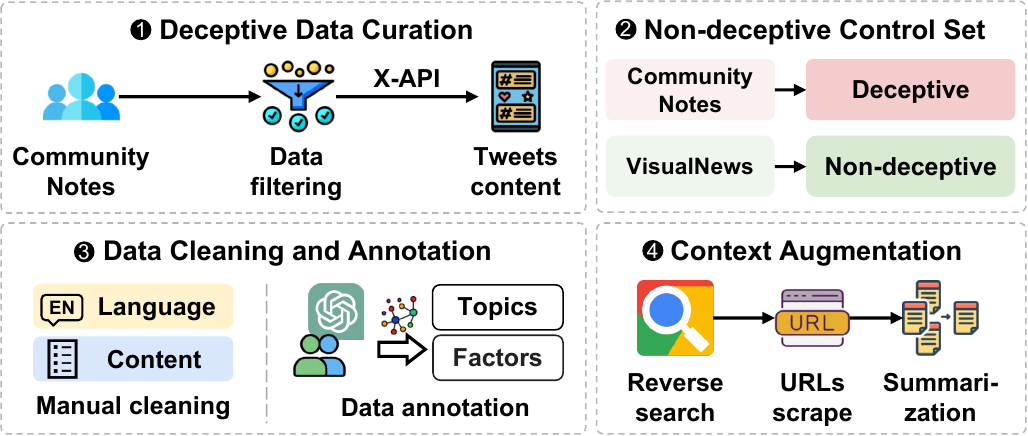}
    \caption{{\dataset} dataset collection pipeline.}
    \label{fig:xnote_pipeline}
\end{figure}

{\dataset} dataset was constructed in four stages as shown in Figure~\ref{fig:xnote_pipeline}: deceptive data curation, non-deceptive control set addition, data cleaning and annotation, and context augmentation via reverse image search.

\subsubsection{Deceptive Data Curation}
We began by downloading the publicly available \textit{Notes} dataset from the X Community Notes program~\cite{xcommunitynote}, which provides note-level metadata for posts flagged as potentially misleading. As shown in Example~1, each community note includes a unique identifier (\datafield{noteId}), the corresponding tweet ID (\datafield{tweetId}), a high-level classification label for indicating misleading or not (\datafield{classification}), binary subtype indicators showing the subcategories of deception (\eg, \datafield{misleading: Missing Important Context}, \datafield{misleading: Outdated Information}), and an explanation written by the contributor (\datafield{summary}). 
Additional metadata includes the media relevance flags (\datafield{isMediaNote}), indicating whether the post is image-based or not. We then identified image-based contextual deception by retaining notes marked with \datafield{isMediaNote=1}, and whose associated posts were tagged \datafield{misleading: Missing Important Context=1}. To exclude content manipulations, we removed records labeled as  \datafield{misleading: Manipulated Media=1}, and we further discarded notes that did not explicitly reference images via a keyword filter (\eg, ``photo'', ``image'', ``picture'', ``photograph''). This procedure yielded an initial set of 3{,}574 notes. Because the \textit{Notes} dataset does not include the tweet content, we used the X-API~\cite{x_api} to retrieve the corresponding tweet data, including the post text, image, timestamp, and other metadata such as retweet count.

\begin{figure}[t]
\begin{tcolorbox}[title=\normalsize Example 1: Original Data in \textit{Notes}, before skip=0pt, after skip=0pt]
\footnotesize
\ttfamily
\begin{verbatim}
id: 1687169266754158592,
text: No *real* election looks like this ...,
date: 2023-08-03 18:31:13,
retweet_count: 8503,
image_urls: [https://pbs.twimg.com/...],
tweet_url: https://twitter.com/...,
community_note: {
  noteId: 1687306040260628480,
  tweetId: 1687169266754158592,
  classification: MISINFORMED_OR_POTENTIALLY_
  MISLEADING,
  misleadingOther: 0,
  misleadingFactualError: 1,
  misleadingManipulatedMedia: 0,
  misleadingOutdatedInformation: 0,
  misleadingMissingImportantContext: 1,
  misleadingUnverifiedClaimAsFact: 0,
  misleadingSatire: 0,
  notMisleadingOther: 0,
  notMisleadingFactuallyCorrect: 0,
  notMisleadingOutdatedButNotWhenWritten: 0,
  notMisleadingClearlySatire: 0,
  notMisleadingPersonalOpinion: 0,
  trustworthySources: 1,
  summary: The image depicts unofficial ...,
  isMediaNote: 1,
  twitter_link: https://twitter.com/...}
\end{verbatim}
\end{tcolorbox}
\vspace{-1.2em}
\end{figure}

\subsubsection{Non-Deceptive Control Set}
To mitigate evaluation bias from a dataset composed solely of deceptive data, we supplemented {\dataset} with a non-deceptive control set sampled from VisualNews~\cite{liu2021visual}. VisualNews contains over one million news images paired with articles, captions, bylines, and metadata from professional news outlets. We randomly selected items from this dataset and treated them as non-deceptive controls. For each selected item, we extracted the image and article text as the dataset content. Publication timestamps are not available in VisualNews; therefore we use reverse search to find the original news article and add the timestamp if applicable. Since these data are from authentic news articles, they do not have accompanying Community Notes. 

\subsubsection{Data Cleaning and Annotation}
After data collection, we conducted a manual cleaning process. Five graduate students were trained using written guidelines and exemplars, and each reviewed a disjoint subset of the 3{,}574 initially collected deceptive candidates. They performed (i) a language check to remove non-English items, and (ii) a content check to verify that each item matches the definition of image-based contextual deception. Items failing either check were removed, leaving 1{,}577 deceptive candidates. Next, two experienced researchers independently re-verified the remaining items. Their agreement was 0.87 measured by Jaccard similarity score. We retained only items labeled as contextual deception by both researchers, resulting in a final English-only set of 1{,}088 posts. To construct a balanced dataset, we then randomly sampled 1{,}088 non-deceptive posts from VisualNews, yielding a final dataset of 2{,}176 items with a balanced class distribution.

After cleaning, we annotated the dataset using a human+LLM collaboration pipeline. We first performed topical categorization with OpenAI GPT-5~\cite{gpt}. GPT-5 was used to propose candidate topics at first, which two researchers consolidated into a 14-topic taxonomy as shown in Table~\ref{tab:topics_only}.
Using each post’s text, image, and Community Note, GPT-5 then assigned one or more topic labels to the data.
Because multiple topics can co-occur in one data entry, we allow multi-label assignments. Then we use the same human+LLM workflow to annotate the fine-grained deceptive factors in the deceptive subset, indicating which factors were misleading in each post. The 10-factor taxonomy used in this process is shown in Table~\ref{tab:factors}. Finally, in a random audit of 100 items, two researchers observed high labeling accuracy at 98\% and 97\%, respectively; we therefore include these LLM-generated annotations in {\dataset} dataset. 

\begin{table}[t]
\centering
\caption{Topic taxonomy in {\dataset} dataset. \textbf{D}: ``deceptive'', \textbf{ND}: ``non-deceptive''. (One data may contain multiple topical labels, so the percentage summation may exceed 100\%.)}
\resizebox{\linewidth}{!}{
\begin{tabular}{rlrrr}
\hline
\textbf{ID} & \textbf{Topic} & \textbf{D} & \textbf{ND} & \textbf{Total} \\
\hline
1  & Politics and Governance & 34.2\% & 23.3\% & 28.7\%\\
2  & War, Conflict, and Geopolitics & 29.2\% & 8.6\% & 18.9\% \\
3  & Conspiracy Narratives & 21.1\% & 0.1\% & 10.6\% \\
4  & Society, Culture, and Identity & 16.3\% & 13.7\% & 15.0\% \\
5  & Science and Technology & 11.6\% & 6.3\% & 9.0\% \\
6  & History and Historical Revisionism & 11.5\% & 3.2\% & 7.4\% \\
7  & Crime, Fraud, and Personal Safety & 10.9\% & 8.4\% & 9.6\% \\
8 & Entertainment and Media & 8.8\% & 30.7\% & 19.8\% \\
9  & Climate, Environment and Disasters & 4.9\% & 6.2\% & 5.5\% \\
10 & Health and Medicine & 4.5\% & 3.0\% & 3.8\% \\
11  & Business, Brands, and Products & 3.0\% & 8.6\% & 5.8\% \\
12 & Public Safety and Emergency Alerts & 2.6\% & 6.8\% & 4.7\% \\
13 & Economy, Finance, and Crypto & 1.8\% & 3.5\% & 2.7\% \\
14 & Others & 0.4\% & 6.1\% & 3.2\% \\
\hline
\end{tabular}
}
\label{tab:topics_only}
\end{table}

\begin{table}[t]
\centering
\caption{Deceptive factor taxonomy in {\dataset} dataset.}
\resizebox{\linewidth}{!}{
\begin{tabular}{rllr}
\hline
\textbf{ID} & \textbf{Factor} & \textbf{Explanation} & \textbf{Percent} \\
\hline
1  & Event & Which event is depicted & 69.0\% \\
2  & Time & When the image was taken & 56.9\% \\
3  & Entity & Who/Which group is involved & 56.6\% \\
4  & Location & Where it occurred & 41.2\% \\
5  & Attribution & Why/who is responsible & 40.4\% \\
6  & Object & What key objects are & 12.1\% \\
7  & Action & What activity/behavior is happening & 12.0\% \\
8  & ImpactScale & How severe/large it is & 5.6\% \\
9  & QuoteSource & Who said it & 2.6\% \\
10 & Others & Any factor not covered & 1.4\% \\
\hline
\end{tabular}
}
\label{tab:factors}
\end{table}

\begin{figure*}[t]
\centering
\begin{minipage}[t]{0.222\textwidth}
  \centering
  \includegraphics[width=\linewidth]{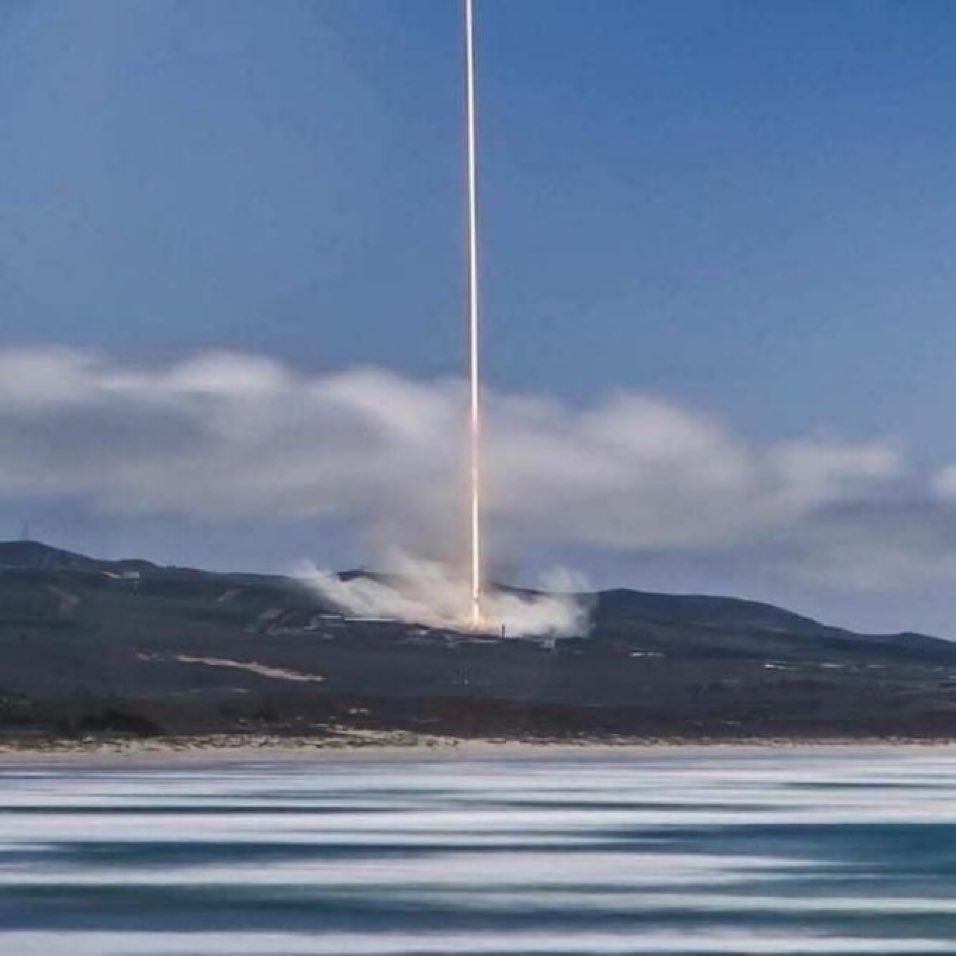}
\end{minipage}
\hfill
\begin{minipage}[t]{0.76\textwidth}
  \begin{tcolorbox}[title=\normalsize Post Metadata, before skip=0pt, after skip=0pt]
    \footnotesize
    \begin{tabularx}{\linewidth}{@{}lX@{}}
      \textbf{Text:}  & This photo is circulating social media. Apparently this beam was captured before the Hawaii fires. Can anyone confirm? \\
      \textbf{Date:}  & 2023-08-10 19:28:57 \\
      \textbf{Retweet \#:} & 257 \\
      \textbf{Topics:}& Public Safety and Emergency Alerts, Science and Technology, Conspiracy Narratives \\
      \textbf{Factors:}& Time, Location, Event \\
      \textbf{Note:}  & The image dates to 2019 and is of a SpaceX Falcon 9 launch from Vandenberg Air force base. \\ 
       & https://arstechnica.com/science/2019/04/rocket-report-darpa-picks-three-aerojet-short-seller-starlink-launch-date/ \\    
       & https://twitter.com/Americanlll/status/1689727802457591809?s=20 \\
    \end{tabularx}
  \end{tcolorbox}
\end{minipage}
\vspace{0.4em}
\begin{tcolorbox}[title=\normalsize External Context, before skip=0pt, after skip=6pt]
\footnotesize
\textbf{URL:} https://newschecker.in/fact-check/hawaii-wildfires-caused-by-laser-strikes-no-viral-images-are-of-unrelated-events \\
\textbf{Summary:} Viral images are of unrelated events from 2018. One image is most likely of a cold-weather phenomenon after a meteor strike in Michigan. The other is of a SpaceX rocket launch in the same year. \\
\textbf{URL:} https://www.thequint.com/news/webqoof/old-unrelated-visuals-shared\ldots
\end{tcolorbox}
\vspace{-1.5em}
\caption{Example data entry in {\dataset}, with image, structured post metadata and external context.}
\label{fig:xnote_example_card}
\end{figure*}

\subsubsection{Context Augmentation}
To support the research for combating contextual deception, we augmented each item with external context through reverse image search, a standard approach for provenance tracing~\cite{qi2024sniffer}. For each image, we used the Google Cloud Vision API~\cite{google_vision} for reverse search, and retained URLs that were marked as \datafield{pages with matching images}. We then fetched each webpage using web loaders from LangChain Community~\cite{langchain}, extracting the title, meta description, and content text. Since the source pages can be lengthy, we generated one summary for each context item using PEGASUS, a news-trained summarizer~\cite{zhang2020pegasus}. Using this procedure, each post was enriched with up to ten external context items, each consisting of a source URL and a summary. Because some images had no web duplicates, some links were unavailable, and API access occasionally failed, we ultimately attached context to 971 deceptive items and 779 non-deceptive items. In total, these 1{,}750 data items with context cover $\sim$80\% of the total 2{,}176 data items in {\dataset}. Table~\ref{tab:data_num} reports the numbers of data in {\dataset} during the dataset collection process, and Figure~\ref{fig:xnote_example_card} shows an example data entry from the final dataset.

\begin{table}[]
    \centering
    \caption{Numbers of remaining data in {\dataset} during the dataset collection process.}
    \label{tab:data_num}
    \resizebox{\linewidth}{!}{
    \begin{tabular}{|c|ccc|c|}
    \hline
       \textbf{Classes} & \textbf{Curation} & \textbf{Cleaning} & \textbf{Final Data} & \textbf{Context} \\
       \hline
       Deceptive & 3{,}574 & 1{,}577 & 1{,}088 & 971 \\
       Non-deceptive & N/A & N/A & 1{,088} & 779 \\
       \hline
       Total & & & 2{,}176 & 1{,}750 \\
       \hline
    \end{tabular}
    }
\end{table}

\subsection{Dataset Analysis}\label{sec:analysis}





\subsubsection{Topic Analysis} 

Table~\ref{tab:topics_only} summarizes the topical taxonomy and label distribution in {\dataset}. Because posts may receive multiple topic labels, the percentages across topics can exceed 100\%. Among deceptive posts, seven topics (IDs 1-7) each account for more than 10\% of data items, showing a relatively diverse distribution. Among these, topic ``1: Politics and Governance'' is the most prevalent. In contrast, non-deceptive posts exhibit a more concentrated distribution. Only three topics exceed 10\% coverage, with topic ``8: Entertainment and Media'' being the most dominant.

\begin{figure}[t]
    \centering
    \includegraphics[width=.95\linewidth]{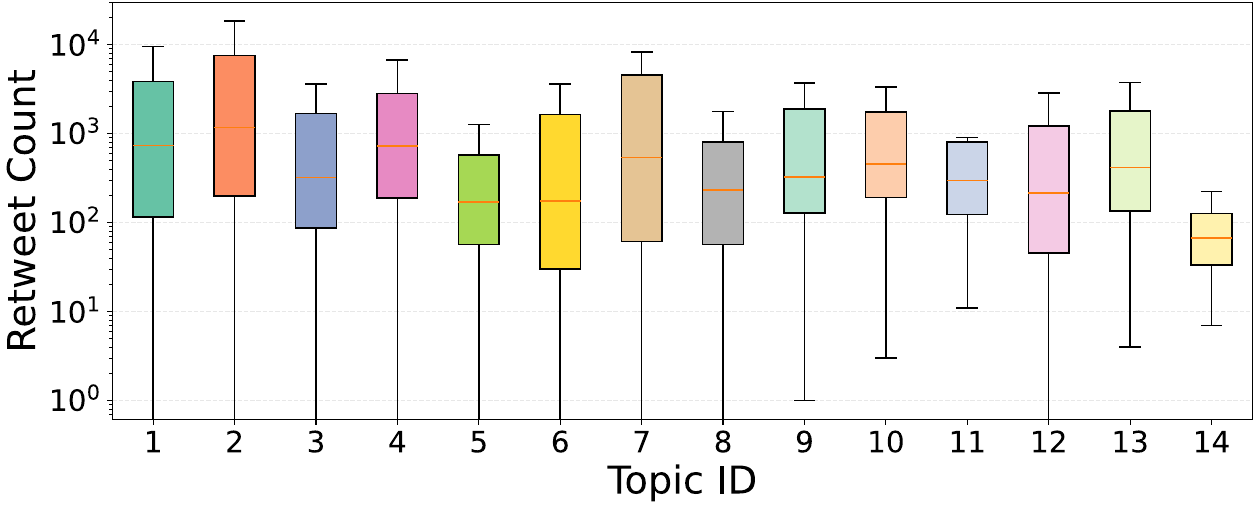}
    \caption{Retweet count distribution related to topics.}
    \label{fig:topic_bar}
\end{figure}

We further analyze retweet engagement and temporal patterns across topics within the deceptive subset. Figure~\ref{fig:topic_bar} shows the retweet count distribution of all 14 topics on a logarithmic scale. Among the seven most prevalent topics, topic ``2: War, Conflict, and Geopolitics'' and topic ``4: Society, Culture, and Identity'' receive the highest retweet engagement, followed by topic ``1: Politics and Governance'' and topic ``7: Crime, Fraud, and Personal Safety.'' We hypothesize that the higher engagement of these topics is related to their emotionally polarizing content, which often amplifies tensions between social groups such as nations, political affiliations, or racial identities. Importantly, higher topic prevalence does not necessarily correspond to higher engagement. For example, topic ``5: Science and Technology'' ranks fifth in terms of post number but exhibits relatively low retweet activity. This suggests limited user interest in propagating deceptive content in this domain. One possible explanation is that deceptive narratives related to science and technology, such as \textit{Flat Earth} claims, are more easily identified and dismissed by users with stronger scientific literacy.

\begin{figure}[t]
    \includegraphics[width=\linewidth]{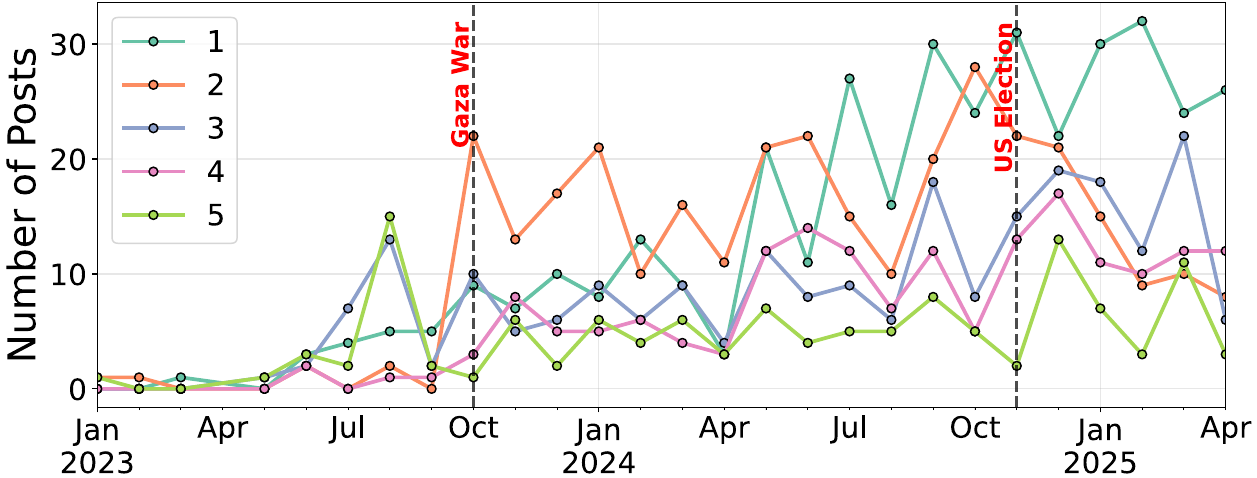}
    \caption{Number of posts trend over time for top-5 topics.}
    \label{fig:topic_trend}
\end{figure}

Figure~\ref{fig:topic_trend} presents the monthly volume trends for the five most prevalent deceptive topics. In addition to an overall increasing trend, which partially reflects the expanded coverage of Community Notes, we observe distinct spikes aligned with major real-world events. For instance, topic ``2: War, Conflict, and Geopolitics'' shows a sharp increase beginning in October 2023, coinciding with the onset of the \textit{Gaza War}. Similarly, topic ``1: Politics and Governance'' increases around July 2024 with the start of the US election campaign and peaks in November during the \textit{US election} month.

\subsubsection{Factor Analysis}
Table~\ref{tab:factors} presents the deceptive factor taxonomy and the distribution of ten factors within the deceptive subset. The results show that deceptive posts frequently manipulate context factors that are not directly observable from the image itself. Specifically, factors such as event, time, entity, location, and attribution each occur in more than 40\% of deceptive posts, suggesting that these factors are the primary targets of contextual deception. In contrast, factors such as object, action, impact scale, and quote source occur substantially less often. These factors typically correspond to visual elements that can be directly inferred from the image content and are therefore easier for users to verify. Overall, this distribution suggests that deceptive content preferentially exploits factors that require external context beyond the image, enabling misleading narratives without altering the visual evidence.

\subsubsection{Source URLs Analysis} 
\begin{figure}[t]
    \centering
    \begin{subfigure}[b]{0.49\linewidth}
        \includegraphics[width=\textwidth]{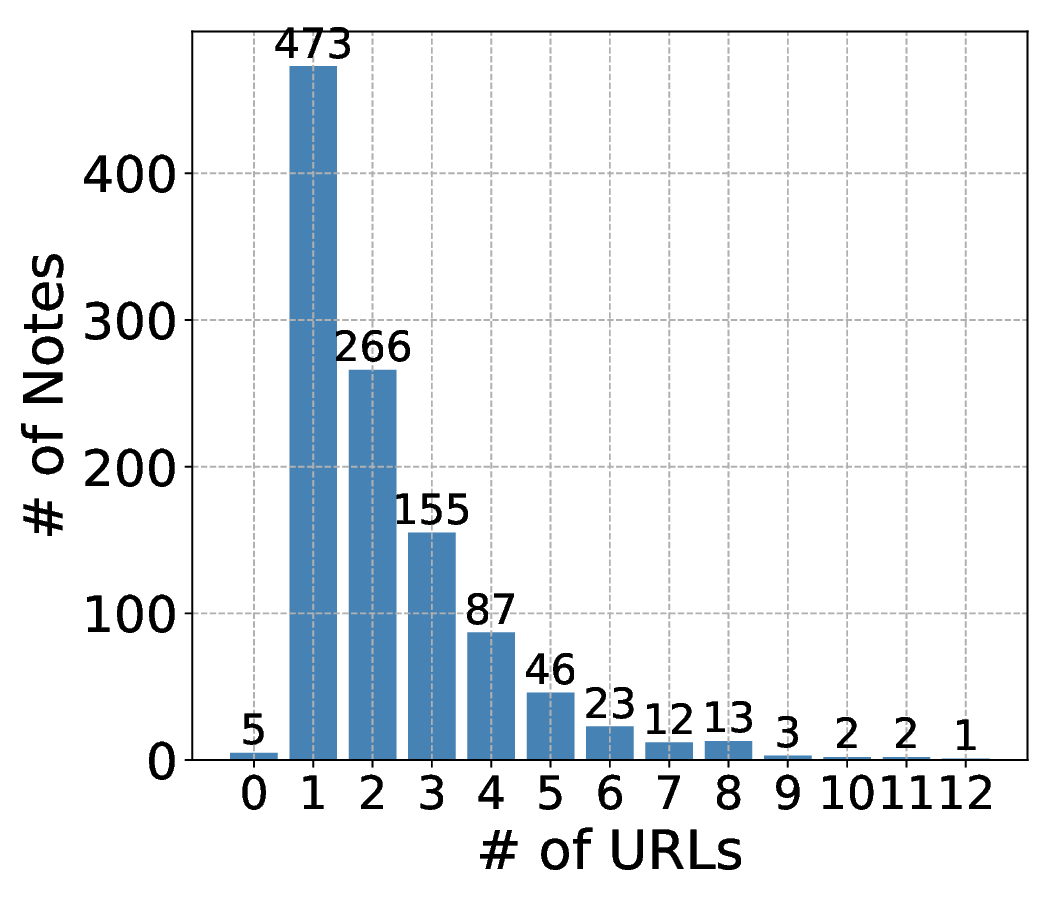}
        \caption{Distribution of URL counts.}
        \label{fig:url_counts}
    \end{subfigure}
    \hfill
    \begin{subfigure}[b]{0.49\linewidth}
        \includegraphics[width=\textwidth]{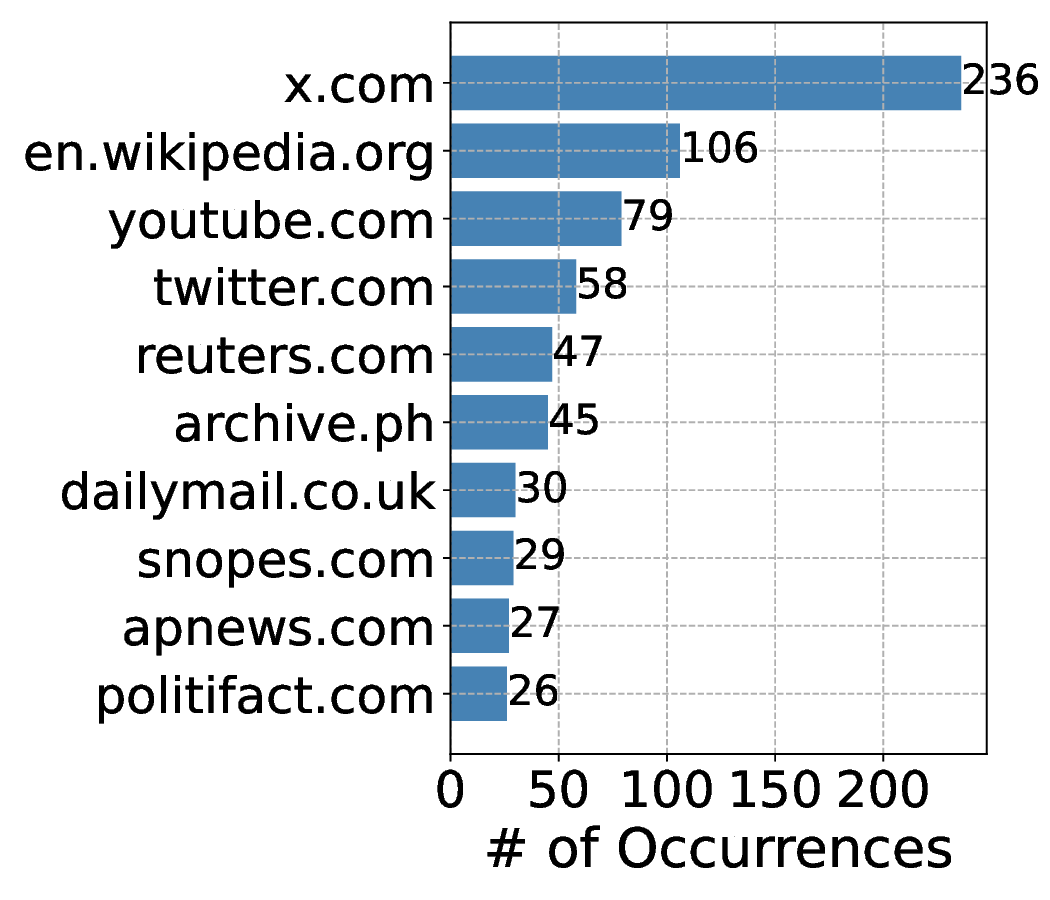}
        \caption{Top-10 domain names.}
        \label{fig:url_domains}
    \end{subfigure}
    \caption{Source URLs analysis in {\dataset}.}
    \label{fig:note_urls}
\end{figure}

We count the number of URLs in each Community Note, and show the distribution of URL counts in Figure~\ref{fig:url_counts}. As shown in the figure, almost all Community Notes contain at least one URL as reference, with only 5 exceptions. This highlights the importance of including URLs in Community Notes~\cite{xcommunitynote,solovev2025references}. 
In addition, we calculate the occurrences of different domain names within all URLs, and show the top-10 domain names in Figure~\ref{fig:url_domains}. As shown in the figure, social media platforms (\eg, x/twitter.com, youtube.com) are the major references in Community Notes, higher than trustworthy news resources such as Wikipedia and news outlets (\eg, reuters.com, dailymail.co.uk, apnews.com). This likely reflects the large volume of unofficial information shared on social platforms rather than by news outlets. However, citing social posts carries reliability risks, as these platforms are also major venues for deception. Archival services (\eg, archive.ph) are also valuable resources, since they help preserve volatile content and provide durable citations when original pages are altered or removed. Surprisingly, fact-checking websites (\eg, snopes.com, politifact.com) are not frequently cited in Community Notes. We attribute this to the inability of fact-checking organizations to keep pace with the scale and speed of online deception. As a result, by the time a Community Note is composed, relevant fact-checking sources are often unavailable for citation.

Our statistical analysis shows that {\dataset} covers diverse topics and deceptive factors, and the Community Notes reference a wide variety of source URLs across domains, positioning it as a key resource for combating image-based contextual deception.


\section{Evaluation}\label{sec:evaluation}
In this section, we benchmark different baselines on {\dataset} for two tasks: deception detection and automated note generation.

\begin{table*}[t]
    \centering
    \caption{Deception detection and note generation results of different methods and  LVLM models. Acc: accuracy; Pre: precision; Rec: recall; R-L: ROUGE-L; MET: METEOR. For each  metric, the best-performing result is highlighted in \colorbox{green!20}{green}.}
    \label{tab:all_results}
    \resizebox{0.88\linewidth}{!}{
    \begin{tabular}{|c|c|c||c|c|c|c||c|c|c|}
    \hline
        \multirow{2}{*}{\textbf{Models}} & \multirow{2}{*}{\textbf{\# Params}} & \multirow{2}{*}{\textbf{Strategies}} & \multicolumn{4}{c||}{\textbf{Detection}} & \multicolumn{3}{c|}{\textbf{Generation}}  \\
        \cline{4-10}
         &  &  & \textbf{F1} & \textbf{Acc} & \textbf{Pre} & \textbf{Rec} & \textbf{BLEU} & \textbf{R-L} & \textbf{MET} \\
        \hline
        \multirow{2}{*}{SNIFFER~\cite{qi2024sniffer}} & \multirow{2}{*}{$\sim$26B} & Closed-book & 0.4202 & 0.5827 & 0.6883 & 0.3024 & 0.0744 & 0.0802 & 0.0603 \\
         &  & Web search & 0.3502 & 0.5634 & 0.6845 & 0.2353 & 0.0795 & 0.0800 & 0.0647  \\
        \hline
        \multirow{2}{*}{Gemma-3~\cite{team2024gemma}} & \multirow{2}{*}{4B} & Closed-book & 0.7525 & 0.7137 & 0.6627 & 0.8704 & 0.2265 & 0.1301 & 0.1348 \\
        &  & Naive RAG & 0.7070 & 0.6797 & 0.6514 & 0.7730 & 0.2299 & 0.1323 & 0.1459 \\
        \hline
        \multirow{2}{*}{InternVL-3.5~\cite{wang2025internvl3_5}} & \multirow{2}{*}{8B} & Closed-book & 0.7118 & 0.6558 & 0.6122 & 0.8502 & 0.2193 & 0.1239 & 0.1302 \\
        &  & Naive RAG & 0.7010 & 0.6374 & 0.5964 & 0.8502 & 0.2234 & 0.1271 & 0.1519 \\
         \hline
        \multirow{2}{*}{LLaVA-OneVision-1.5~\cite{an2025llavaonevision15fullyopenframework}} & \multirow{2}{*}{8B} & Closed-book & 0.7961 & 0.7992 & 0.8085 & 0.7840 & 0.1993 & 0.1084 & 0.1285 \\
        &  & Naive RAG & 0.7443 & 0.7284 & 0.7032 & 0.7904 & 0.2202 & 0.1226 & 0.1555 \\
        \hline
        \multirow{2}{*}{NVILA~\cite{liu2024nvila}} & \multirow{2}{*}{8B} & Closed-book & 0.7564 & 0.7845 & \cellcolor{green!20}0.8698 & 0.6691 & 0.2241 & 0.1216 & 0.1406 \\
        &  & Naive RAG & 0.6670 & 0.7040 & 0.7624 & 0.5928 & 0.2226 & 0.1264 & 0.1523 \\
        \hline
        \multirow{2}{*}{Qwen3-VL~\cite{qwen3technicalreport}} & \multirow{2}{*}{7B} & Closed-book & 0.7319 & 0.6489 & 0.5919 & \cellcolor{green!20}0.9586 & 0.2304 & 0.1295 & 0.1467 \\
        &  & Naive RAG & 0.7135 & 0.6622 & 0.6195 & 0.8410 & \cellcolor{green!20}0.2309 & 0.1345 & \cellcolor{green!20}0.1612  \\
        \hline
         \multirow{2}{*}{GPT-5~\cite{gpt}} & \multirow{2}{*}{-} & Closed-book & \cellcolor{green!20}0.8359 & \cellcolor{green!20}0.8359 & 0.8362 & 0.8355 & 0.2281 & 0.1327 & 0.1439 \\
         &  & Web search & 0.7917 & 0.7960 & 0.8084 & 0.7757 & 0.2259 & \cellcolor{green!20}0.1505 & 0.1448 \\
    \hline
    \end{tabular}
    }
\end{table*}

\textbf{Baselines.} We evaluate five open-source frontier LVLM backbones: Gemma-3~\cite{team2024gemma}, InternVL-3.5~\cite{wang2025internvl3_5}, LLaVA-OneVision-1.5~\cite{an2025llavaonevision15fullyopenframework}, NVILA~\cite{liu2024nvila}, and Qwen3-VL~\cite{qwen3technicalreport}. To ensure computational efficiency, we select lightweight variants with fewer than 10B parameters and use instruction-tuned versions when available. We evaluate each model under two settings: (i) \textit{closed-book} without external context; (ii) \textit{naive RAG} with all retrieved context concatenated into the prompt. For comparison, we include an end-to-end contextual deception detector, SNIFFER~\cite{qi2024sniffer} (InstructBLIP@13B + Vicuna@13B), which could use either \textit{closed-book} setting or \textit{web search} setting (via similar Google reverse search). We also evaluate a lightweight commercial tool, GPT-5~\cite{gpt}. Since GPT5 provides an integrated web search capability, we report results under two configurations: one of \textit{closed-book} and one with \textit{web search} tool. Detailed prompts are included in our public code repository.

\textbf{Evaluation metrics.} For deception detection task, we treat it as binary classification with \textit{deceptive} as the positive class and \textit{non-deceptive} as the negative class, reporting F1, Accuracy, Precision, and Recall. For the automated note generation task, we evaluate generated notes against the ground-truth Community Notes using three widely used metrics, BLEU~\cite{papineni2002bleu}, ROUGE-L~\cite{lin2004rouge}, and METEOR~\cite{banerjee2005meteor}. As ground-truth Community Notes are only available for deceptive data, we restrict evaluation of the note generation task to the deceptive subset.



\subsection{Results Analysis} 
Table~\ref{tab:all_results} reports results across all methods and LVLM backbones, with best results highlighted in green. The end-to-end model SNIFFER consistently underperforms across tasks. This is likely due to its reliance on synthetic training data, which does not capture the complexity of real-world contextual deception in {\dataset}. Among open-source LVLM models, LLaVA-OneVision-1.5 achieves the best detection performance (F1: 0.7961, closed-book), while Qwen3-VL performs best on note generation (ROUGE-L: 0.1345, naive RAG), indicating differing strengths across tasks. The commercial model GPT-5 achieves the strongest detection performance overall and remains competitive on the note generation task, but at a relatively high cost (\$0.03 per API call in our experiments). From our analysis, we identify two key observations that require attention for future research on automated Community Notes generation:
\begin{enumerate}
    \item \textbf{Naive RAG is not consistently reliable.} It often degrades detection performance due to noisy or irrelevant retrieved content~\cite{salemi2024evaluating,chang2024main}, while providing modest gains for generation. This suggests that external context requires more effective filtering and grounding before being incorporated into model inputs.
    \item \textbf{Standard metrics are insufficient.} Overlap-based metrics (BLEU, ROUGE-L, METEOR) fail to capture whether generated notes are truly helpful to users, particularly in terms of credibility (\eg, citing reliable sources). This highlights the need for user-aligned evaluation metrics.
\end{enumerate}

\begin{figure}[t]
    \centering
    \includegraphics[width=0.95\linewidth]{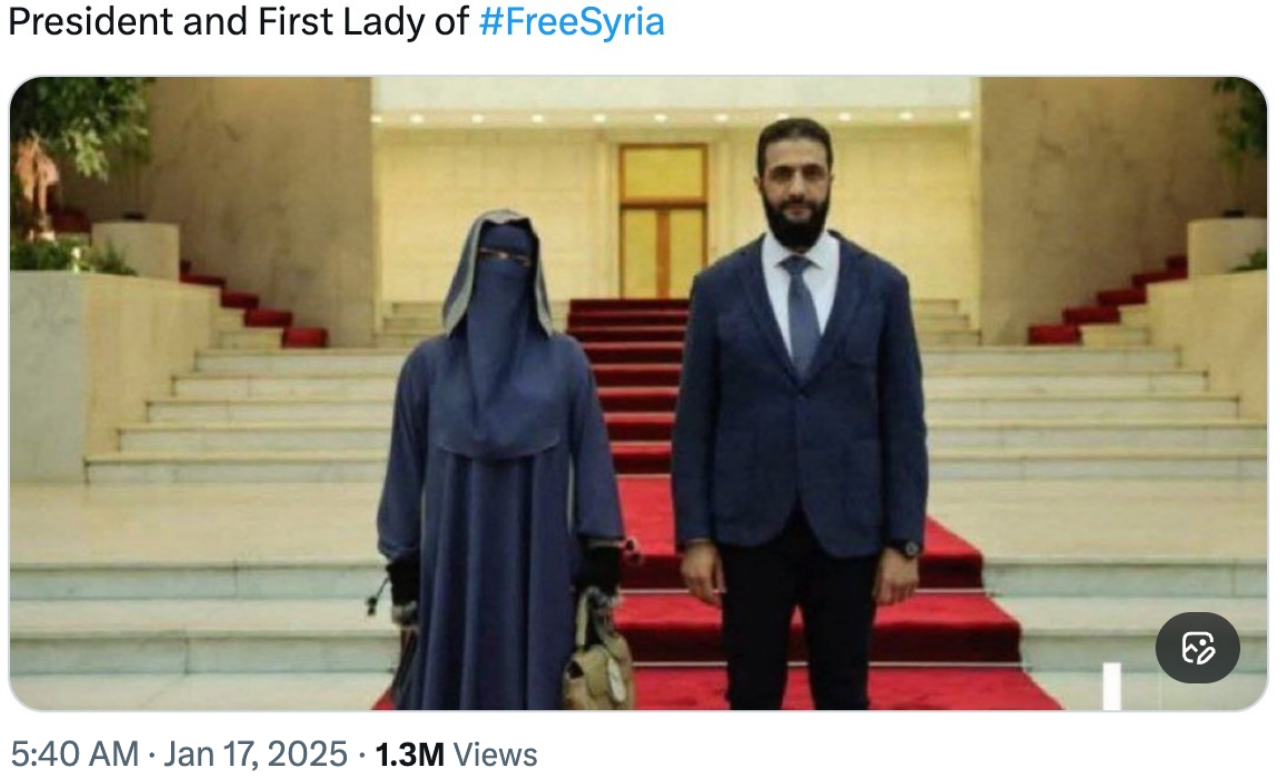}
    \caption{The original X post for Example 2.}
    \label{fig:example2}
\begin{tcolorbox}[title=\normalsize Example 2: Qualitative Example of Different Notes]
    \footnotesize
    \textbf{Community Note:} The photo shows Ahmed Al-Sharaa with the \textcolor{blue}{poet Alaa Mustafa Ali}, who was part of a delegation of Syrian writers and poets who met him in Damascus on January 15, 2025. She is not First Lady of Syria, as some have falsely reported.    https://www.annahar.com/Factcheck/188268/... \\
    \textbf{Qwen3-VL (closed-book):} There is no such entity as the ``Free Syria'' government, and the individuals in the image are not the President and First Lady of Syria. The image shows a man and woman posing in front of a building with red carpeted stairs, but \textcolor{red}{neither is a recognized official of the Syrian government}, and the claim is factually incorrect. \\
    \textbf{Qwen3-VL (naive RAG):} The photo does not show Ahmed al-Sharaa with his wife; instead, it depicts him posing with \textcolor{blue}{a female poet}, as confirmed by fact-checking from Misbar. The claim that they are the President and First Lady of \#FreeSyria is misleading. \\
    \textbf{GPT-5 (web search):} The photo is from a Jan. 15, 2025 meeting with Syrian poets and writers; the woman pictured is \textcolor{blue}{a poet}, not a First Lady, so it is not a ``President and First Lady of Free Syria'' image. ([checkyourfact.com](https://checkyourfact.com/2025/01/24/...)) 
\end{tcolorbox}
\vspace{-1.5em}
\end{figure}

To illustrate differences in generated notes, we present a representative case study. The original X post is shown in Figure~\ref{fig:example2}, with corresponding outputs in Example~2, where correct information is marked as blue text, incorrect content is marked as red text. The post falsely claims that the image depicts the ``President and First Lady of Syria.'' In contrast, the ground-truth Community Note clarifies that the image shows Syria's president Ahmed Al-Sharaa with ``poet Alaa Mustafa Ali'', not the First Lady. Qwen3-VL under the closed-book setting incorrectly claims that ``neither is a recognized official of the Syrian government.'' When equipped with naive RAG, Qwen3-VL correctly identifies the misrepresented entity using external context, but fails to include a credible source URL in its generated note. GPT-5 with web search, on the other hand, generates a correct and concise explanation while also providing a supporting source URL. This example underscores the importance of accurate \textit{contextual understanding}, the inclusion of \textit{credible sources}, and \textit{appropriate metrics} to evaluate these aspects.

\section{Dataset Release and Ethical Considerations}
{\dataset} is constructed from publicly accessible posts on X. We only use content that is available without accessing private accounts or messages. All screenshots and images reproduced in this paper are sourced from X, with the platform credited accordingly. Copyright and ownership of the original media remain with their respective rights holders. To support reproducibility while respecting platform policies and copyright constraints, we release {\dataset} through two complementary channels:

(1) \textbf{Policy-compliant release}. We provide a public GitHub repository containing public code and a policy-compliant version of {\dataset}, which includes only tweet IDs and associated annotations, without redistributing raw media. Reconstructing the full content requires ``hydration'' via the official X API~\cite{x_api} or other compliant procedures. Any hydrated content remains subject to X’s platform policies and the rights of the original content creators. GitHub: \url{https://github.com/MaJinWakeUp/XNote}.

(2) \textbf{Controlled-access release}. We additionally release a version of {\dataset} on Hugging Face under a gated access mechanism. This version is distributed under the Creative Commons Attribution-NonCommercial 4.0 (CC BY-NC 4.0) license. Access is granted upon request for academic and research purposes only; commercial use is strictly prohibited. Hugging Face: \url{https://huggingface.co/datasets/majinwakeup30/XNote}.

\section{Conclusion} 
In this work, we introduce {\dataset}, a real-world multimodal dataset designed to support research on automated Community Notes generation for image-based contextual deception. {\dataset} consists of X posts paired with ground-truth Community Notes, along with retrieved external context and annotations of post topics and manipulation factors. We further benchmark a range of frontier LVLMs on both deception detection and note generation tasks, analyzing their performance under different settings. Our results reveal key challenges in automated Community Notes generation, highlighting the need for more reliable methods and evaluation metrics.



\bibliographystyle{ACM-Reference-Format}
\bibliography{references}

@String{Computing = "Computing" }

@String{Computer = "{IEEE} Computer" }

@String{Springer = "Springer-Verlag" }

@article{allen2024quantifying,
  title={Quantifying the impact of misinformation and vaccine-skeptical content on Facebook},
  author={Allen, Jennifer and Watts, Duncan J and Rand, David G},
  journal={Science},
  volume={384},
  number={6699},
  pages={eadk3451},
  year={2024},
  publisher={American Association for the Advancement of Science}
}

@article{aw2021covid,
  title={COVID-19 vaccine hesitancy—A scoping review of literature in high-income countries},
  author={Aw, Junjie and Seng, Jun Jie Benjamin and Seah, Sharna Si Ying and Low, Lian Leng},
  journal={Vaccines},
  volume={9},
  number={8},
  pages={900},
  year={2021},
  publisher={MDPI}
}

@article{tsikerdekis2014online,
  title={Online deception in social media},
  author={Tsikerdekis, Michail and Zeadally, Sherali},
  journal={Communications of the ACM},
  volume={57},
  number={9},
  pages={72--80},
  year={2014},
  publisher={ACM New York, NY, USA}
}

@article{li2020picture,
  title={Is a picture worth a thousand words? An empirical study of image content and social media engagement},
  author={Li, Yiyi and Xie, Ying},
  journal={Journal of marketing research},
  volume={57},
  number={1},
  pages={1--19},
  year={2020},
  publisher={Sage Publications Sage CA: Los Angeles, CA}
}

@article{abdali2024multi,
  title={Multi-modal misinformation detection: Approaches, challenges and opportunities},
  author={Abdali, Sara and Shaham, Sina and Krishnamachari, Bhaskar},
  journal={ACM Computing Surveys},
  volume={57},
  number={3},
  pages={1--29},
  year={2024},
  publisher={ACM New York, NY}
}

@inproceedings{zhang2020pegasus,
  title={Pegasus: Pre-training with extracted gap-sentences for abstractive summarization},
  author={Zhang, Jingqing and Zhao, Yao and Saleh, Mohammad and Liu, Peter},
  booktitle={International conference on machine learning},
  pages={11328--11339},
  year={2020},
  organization={PMLR}
}

@inproceedings{salemi2024evaluating,
  title={Evaluating retrieval quality in retrieval-augmented generation},
  author={Salemi, Alireza and Zamani, Hamed},
  booktitle={Proceedings of the 47th International ACM SIGIR Conference on Research and Development in Information Retrieval},
  pages={2395--2400},
  year={2024}
}

@article{chang2024main,
  title={Main-rag: Multi-agent filtering retrieval-augmented generation},
  author={Chang, Chia-Yuan and Jiang, Zhimeng and Rakesh, Vineeth and Pan, Menghai and Yeh, Chin-Chia Michael and Wang, Guanchu and Hu, Mingzhi and Xu, Zhichao and Zheng, Yan and Das, Mahashweta and others},
  journal={arXiv preprint arXiv:2501.00332},
  year={2024}
}

@article{liu2024mmfakebench,
  title={Mmfakebench: A mixed-source multimodal misinformation detection benchmark for lvlms},
  author={Liu, Xuannan and Li, Zekun and Li, Peipei and Huang, Huaibo and Xia, Shuhan and Cui, Xing and Huang, Linzhi and Deng, Weihong and He, Zhaofeng},
  journal={arXiv preprint arXiv:2406.08772},
  year={2024}
}

@misc{schemaorg_claimreview,
  author       = {{Schema.org}},
  title        = {ClaimReview — Schema.org Type},
  howpublished = {\url{https://schema.org/ClaimReview}},
  note         = {Accessed: October, 2025}
}

@misc{google_claimreview_structured_data,
  author       = {{Google Search Central}},
  title        = {Fact check (ClaimReview) structured data},
  howpublished = {\url{https://developers.google.com/search/docs/appearance/structured-data/factcheck}},
  note         = {Accessed: October, 2025}
}

@article{hu2025overview,
  title={An overview of fake news detection: From a new perspective},
  author={Hu, Bo and Mao, Zhendong and Zhang, Yongdong},
  journal={Fundamental Research},
  volume={5},
  number={1},
  pages={332--346},
  year={2025},
  publisher={Elsevier}
}

@article{radford2021learning,
  title        = {Learning Transferable Visual Models From Natural Language Supervision},
  author       = {Radford, Alec and Kim, Jong Wook and Hallacy, Chris and Ramesh, Aditya and Goh, Gabriel and Agarwal, Sandhini and Sastry, Girish and Askell, Amanda and Mishkin, Pamela and Clark, Jack and Krueger, Gretchen and Sutskever, Ilya},
  journal      = {arXiv preprint arXiv:2103.00020},
  year         = {2021}
}

@inproceedings{de2025supernotes,
  title={Supernotes: Driving consensus in crowd-sourced fact-checking},
  author={De, Soham and Bakker, Michiel A and Baxter, Jay and Saveski, Martin},
  booktitle={Proceedings of the ACM on Web Conference 2025},
  pages={3751--3761},
  year={2025}
}

@article{mirsky2021creation,
  title={The creation and detection of deepfakes: A survey},
  author={Mirsky, Yisroel and Lee, Wenke},
  journal={ACM computing surveys (CSUR)},
  volume={54},
  number={1},
  pages={1--41},
  year={2021},
  publisher={ACM New York, NY, USA}
}

@article{westerlund2019emergence,
  title={The emergence of deepfake technology: A review},
  author={Westerlund, Mika},
  journal={Technology innovation management review},
  volume={9},
  number={11},
  year={2019}
}

@article{yuan2023support,
  title={Support or refute: Analyzing the stance of evidence to detect out-of-context mis-and disinformation},
  author={Yuan, Xin and Guo, Jie and Qiu, Weidong and Huang, Zheng and Li, Shujun},
  journal={arXiv preprint arXiv:2311.01766},
  year={2023}
}

@article{lin2024detecting,
  title={Detecting multimedia generated by large ai models: A survey},
  author={Lin, Li and Gupta, Neeraj and Zhang, Yue and Ren, Hainan and Liu, Chun-Hao and Ding, Feng and Wang, Xin and Li, Xin and Verdoliva, Luisa and Hu, Shu},
  journal={arXiv preprint arXiv:2402.00045},
  year={2024}
}

@inproceedings{wang2021understanding,
  title={Understanding the use of fauxtography on social media},
  author={Wang, Yuping and Tahmasbi, Fatemeh and Blackburn, Jeremy and Bradlyn, Barry and De Cristofaro, Emiliano and Magerman, David and Zannettou, Savvas and Stringhini, Gianluca},
  booktitle={Proceedings of the International AAAI Conference on Web and Social Media},
  volume={15},
  pages={776--786},
  year={2021}
}

@inproceedings{reimers2019sentence,
  title        = {Sentence-{BERT}: Sentence Embeddings using Siamese BERT-Networks},
  author       = {Reimers, Nils and Gurevych, Iryna},
  booktitle    = {Proceedings of the 2019 Conference on Empirical Methods in Natural Language Processing},
  pages        = {3982--3992},
  year         = {2019},
  doi          = {10.18653/v1/D19-1410}
}

@article{papadopoulos2024verite,
  title={Verite: a robust benchmark for multimodal misinformation detection accounting for unimodal bias},
  author={Papadopoulos, Stefanos-Iordanis and Koutlis, Christos and Papadopoulos, Symeon and Petrantonakis, Panagiotis C},
  journal={International Journal of Multimedia Information Retrieval},
  volume={13},
  number={1},
  pages={4},
  year={2024},
  publisher={Springer}
}

@article{luo2021newsclippings,
  title={Newsclippings: Automatic generation of out-of-context multimodal media},
  author={Luo, Grace and Darrell, Trevor and Rohrbach, Anna},
  journal={arXiv preprint arXiv:2104.05893},
  year={2021}
}

@inproceedings{xu2024mmooc,
  title={MMOOC: A Multimodal Misinformation Dataset for Out-of-Context News Analysis},
  author={Xu, Qingzheng and Du, Heming and Chen, Huiqiang and Liu, Bo and Yu, Xin},
  booktitle={Australasian Conference on Information Security and Privacy},
  pages={444--459},
  year={2024},
  organization={Springer}
}

@inproceedings{liu2021visual,
  title={Visual News: Benchmark and Challenges in News Image Captioning},
  author={Liu, Fuxiao and Wang, Yinghan and Wang, Tianlu and Ordonez, Vicente},
  booktitle={Proceedings of the 2021 Conference on Empirical Methods in Natural Language Processing},
  pages={6761--6771},
  year={2021}
}

@article{pennycook2021psychology,
  title={The psychology of fake news},
  author={Pennycook, Gordon and Rand, David G},
  journal={Trends in cognitive sciences},
  volume={25},
  number={5},
  pages={388--402},
  year={2021},
  publisher={Elsevier}
}

@article{lewandowsky2023misinformation,
  title={Misinformation and the epistemic integrity of democracy},
  author={Lewandowsky, Stephan and Ecker, Ullrich KH and Cook, John and Van Der Linden, Sander and Roozenbeek, Jon and Oreskes, Naomi},
  journal={Current opinion in psychology},
  volume={54},
  pages={101711},
  year={2023},
  publisher={Elsevier}
}

@misc{photoshop,
  author = "{Adobe}",
  title = {Photoshop},
  howpublished = {\url{https://www.adobe.com/products/photoshop}},
  year = {2025},
  note = {Accessed: September, 2025}
}

@misc{xcommunitynote,
  author = "{X Corp.}",
  title = {X Community Notes},
  howpublished = {\url{https://communitynotes.x.com/guide/en/about/introduction}},
  year = {2025},
  note = {Accessed: September, 2025}
}

@misc{metacommunitynote,
  author = "{Meta}",
  title = {Introducing Community Notes},
  howpublished = {\url{https://www.meta.com/technologies/community-notes/}},
  year = {2025},
  note = {Accessed: September, 2025}
}

@inproceedings{qi2024sniffer,
  title={Sniffer: Multimodal large language model for explainable out-of-context misinformation detection},
  author={Qi, Peng and Yan, Zehong and Hsu, Wynne and Lee, Mong Li},
  booktitle={Proceedings of the IEEE/CVF conference on computer vision and pattern recognition},
  pages={13052--13062},
  year={2024}
}

@article{dufour2024ammeba,
  title={Ammeba: A large-scale survey and dataset of media-based misinformation in-the-wild},
  author={Dufour, Nicholas and Pathak, Arkanath and Samangouei, Pouya and Hariri, Nikki and Deshetti, Shashi and Dudfield, Andrew and Guess, Christopher and Escayola, Pablo Hern{\'a}ndez and Tran, Bobby and Babakar, Mevan and others},
  journal={arXiv preprint arXiv:2405.11697},
  year={2024}
}

@misc{x_api,
  title = {X Developer Platform API},
  author = {{X Corp.}},
  year = {2025},
  howpublished = {\url{https://developer.x.com/en/portal/dashboard}},
  note = {Accessed: April, 2025}
}

@misc{politifact,
  title = {PolitiFact},
  author = {{The Poynter Institute}},
  year = {2026},
  howpublished = {\url{https://www.politifact.com/}},
  note = {Accessed: January, 2026}
}

@misc{snopes,
  title = {Snopes},
  author = {{Snopes, Inc.}},
  year = {2026},
  howpublished = {\url{https://www.snopes.com/}},
  note = {Accessed: January, 2026}
}

@misc{factcheck,
  title = {FactCheck},
  author = {{FactCheck.org}},
  year = {2025},
  howpublished = {\url{https://www.factcheck.org/}},
  note = {Accessed: January, 2026}
}

@article{aimeur2023fake,
  title={Fake news, disinformation and misinformation in social media: a review},
  author={A{\"\i}meur, Esma and Amri, Sabrine and Brassard, Gilles},
  journal={Social Network Analysis and Mining},
  volume={13},
  number={1},
  pages={30},
  year={2023},
  publisher={Springer}
}

@misc{google_vision,
  author = {{Google Cloud}},
  title = {Detect Web Entities and Pages},
  howpublished = {\url{https://cloud.google.com/vision/docs/detecting-web}},
  note = {Accessed: April, 2025},
  year = {2025}
}

@misc{gpt,
  author = {{OpenAI}},
  title = {OpenAI Models},
  howpublished = {\url{https://platform.openai.com/docs/models/}},
  note = {Accessed: September, 2025},
  year = {2025}
}

@misc{langchain,
  author = {{LangChain}},
  title = {LangChain Community},
  howpublished = {\url{https://python.langchain.com/api_reference/community/index.html}},
  note = {Accessed: June, 2025},
  year = {2025}
}

@article{liu2024nvila,
  title={NVILA: Efficient frontier visual language models},
  author={Liu, Zhijian and Zhu, Ligeng and Shi, Baifeng and Zhang, Zhuoyang and Lou, Yuming and Yang, Shang and Xi, Haocheng and Cao, Shiyi and Gu, Yuxian and Li, Dacheng and others},
  journal={arXiv preprint arXiv:2412.04468},
  year={2024}
}

@inproceedings{papineni2002bleu,
  title={Bleu: a method for automatic evaluation of machine translation},
  author={Papineni, Kishore and Roukos, Salim and Ward, Todd and Zhu, Wei-Jing},
  booktitle={Proceedings of the 40th annual meeting of the Association for Computational Linguistics},
  pages={311--318},
  year={2002}
}

@inproceedings{lin2004rouge,
  title={Rouge: A package for automatic evaluation of summaries},
  author={Lin, Chin-Yew},
  booktitle={Text summarization branches out},
  pages={74--81},
  year={2004}
}

@inproceedings{banerjee2005meteor,
  title={METEOR: An automatic metric for MT evaluation with improved correlation with human judgments},
  author={Banerjee, Satanjeev and Lavie, Alon},
  booktitle={Proceedings of the acl workshop on intrinsic and extrinsic evaluation measures for machine translation and/or summarization},
  pages={65--72},
  year={2005}
}

@article{yang2023visual,
  title={Visual misinformation on Facebook},
  author={Yang, Yunkang and Davis, Trevor and Hindman, Matthew},
  journal={Journal of Communication},
  volume={73},
  number={4},
  pages={316--328},
  year={2023},
  publisher={Oxford University Press}
}

@article{nakamura2019fakeddit,
  title={r/fakeddit: A new multimodal benchmark dataset for fine-grained fake news detection},
  author={Nakamura, Kai and Levy, Sharon and Wang, William Yang},
  journal={arXiv preprint arXiv:1911.03854},
  year={2019}
}

@inproceedings{rangapur2025fin,
  title={Fin-Fact: A Benchmark Dataset for Multimodal Financial Fact-Checking and Explanation Generation},
  author={Rangapur, Aman and Wang, Haoran and Jian, Ling and Shu, Kai},
  booktitle={Companion Proceedings of the ACM on Web Conference 2025},
  pages={785--788},
  year={2025}
}

@article{solovev2025references,
  title={References to unbiased sources increase the helpfulness of community fact-checks},
  author={Solovev, Kirill and Pr{\"o}llochs, Nicolas},
  journal={Scientific Reports},
  volume={15},
  number={1},
  pages={25749},
  year={2025},
  publisher={Nature Publishing Group UK London}
}

@article{team2024gemma,
  title={Gemma: Open models based on gemini research and technology},
  author={Team, Gemma and Mesnard, Thomas and Hardin, Cassidy and Dadashi, Robert and Bhupatiraju, Surya and Pathak, Shreya and Sifre, Laurent and Rivi{\`e}re, Morgane and Kale, Mihir Sanjay and Love, Juliette and others},
  journal={arXiv preprint arXiv:2403.08295},
  year={2024}
}

@article{wang2025internvl3_5,
  title={InternVL3.5: Advancing Open-Source Multimodal Models in Versatility, Reasoning, and Efficiency},
  author={Wang, Weiyun and Gao, Zhangwei and Gu, Lixin and Pu, Hengjun and Cui, Long and Wei, Xingguang and Liu, Zhaoyang and Jing, Linglin and Ye, Shenglong and Shao, Jie and others},
  journal={arXiv preprint arXiv:2508.18265},
  year={2025}
}

@misc{an2025llavaonevision15fullyopenframework,
      title={LLaVA-OneVision-1.5: Fully Open Framework for Democratized Multimodal Training}, 
      author={Xiang An and Yin Xie and Kaicheng Yang and Wenkang Zhang and Xiuwei Zhao and Zheng Cheng and Yirui Wang and Songcen Xu and Changrui Chen and Chunsheng Wu and Huajie Tan and Chunyuan Li and Jing Yang and Jie Yu and Xiyao Wang and Bin Qin and Yumeng Wang and Zizhen Yan and Ziyong Feng and Ziwei Liu and Bo Li and Jiankang Deng},
      year={2025},
      eprint={2509.23661},
      archivePrefix={arXiv},
      primaryClass={cs.CV},
      url={https://arxiv.org/abs/2509.23661}, 
}

@misc{qwen3technicalreport,
      title={Qwen3 Technical Report}, 
      author={Qwen Team},
      year={2025},
      eprint={2505.09388},
      archivePrefix={arXiv},
      primaryClass={cs.CL},
      url={https://arxiv.org/abs/2505.09388}, 
}


\end{document}